\documentclass[lettersize,journal]{IEEEtran}
\usepackage{amsmath,amsfonts}
\usepackage{algorithmic}
\usepackage{algorithm}
\usepackage{array}
\usepackage[caption=false,font=normalsize,labelfont=sf,textfont=sf]{subfig}
\usepackage{textcomp}
\usepackage{stfloats}
\usepackage{url}
\usepackage{verbatim}
\usepackage{graphicx}
\usepackage{cite}
\usepackage{multirow}
\usepackage{float}
\usepackage{booktabs}
\usepackage[colorlinks,linkcolor=red,anchorcolor=blue,citecolor=green]{hyperref}
\usepackage[switch]{lineno}

\begin{document}

\title{FusePose: IMU-Vision Sensor Fusion in Kinematic Space for Parametric Human Pose Estimation}

\author{
Yiming Bao, Xu Zhao*, \IEEEmembership{Member, IEEE}, Dahong Qian*, \IEEEmembership{Senior Member, IEEE}

\thanks{X. Zhao and D. Qian are co-corresponding authors.}

\thanks{Y. Bao and D. Qian are with the School of Biomedical Engineering, Shanghai Jiao Tong University, Shanghai, China (e-mail: yiming.bao@sjtu.edu.cn; dahong.qian@sjtu.edu.cn)}

\thanks{X. Zhao is with the Department of Automation, Shanghai Jiao Tong University, Shanghai, China (e-mail: zhaoxu@sjtu.edu.cn)}

\thanks{This work has been supported by the NSFC grants 62176156 and Deepwise Healthcare Joint Research Lab, Shanghai Jiao Tong University.}
}
\markboth{Journal of \LaTeX\ Class Files,~Vol.~14, No.~8, August~2021}%
{Shell \MakeLowercase{\textit{Bao et al.}}: A Sample Article Using IEEEtran.cls for IEEE Journals}


\maketitle

\begin{abstract}
There exist challenging problems in 3D human pose estimation mission, such as poor performance caused by occlusion and self-occlusion. Recently, IMU-vision sensor fusion is regarded as valuable for solving these problems. However, previous researches on the fusion of IMU and vision data, which is heterogeneous, fail to adequately utilize either IMU raw data or reliable high-level vision features. To facilitate a more efficient sensor fusion, in this work we propose a framework called \emph{FusePose} under a parametric human kinematic model. Specifically, we aggregate different information of IMU or vision data and introduce three distinctive sensor fusion approaches: NaiveFuse, KineFuse and AdaDeepFuse. NaiveFuse servers as a basic approach that only fuses simplified IMU data and estimated 3D pose in euclidean space. While in kinematic space, KineFuse is able to integrate the calibrated and aligned IMU raw data with converted 3D pose parameters. AdaDeepFuse further develops this kinematical fusion process to an adaptive and end-to-end trainable manner. Comprehensive experiments with ablation studies demonstrate the rationality and superiority of the proposed framework. The performance of 3D human pose estimation is improved compared to the baseline result. On Total Capture dataset, KineFuse surpasses previous state-of-the-art which uses IMU only for testing by 8.6\%. AdaDeepFuse surpasses state-of-the-art which uses IMU for both training and testing by 8.5\%. Moreover, we validate the generalization capability of our framework through experiments on Human3.6M dataset.
\end{abstract}


\section{Introduction}
\label{sec:intro}

\begin{figure}[t]
 \begin{center}
 	\centerline{\includegraphics[width= 8.5cm ]{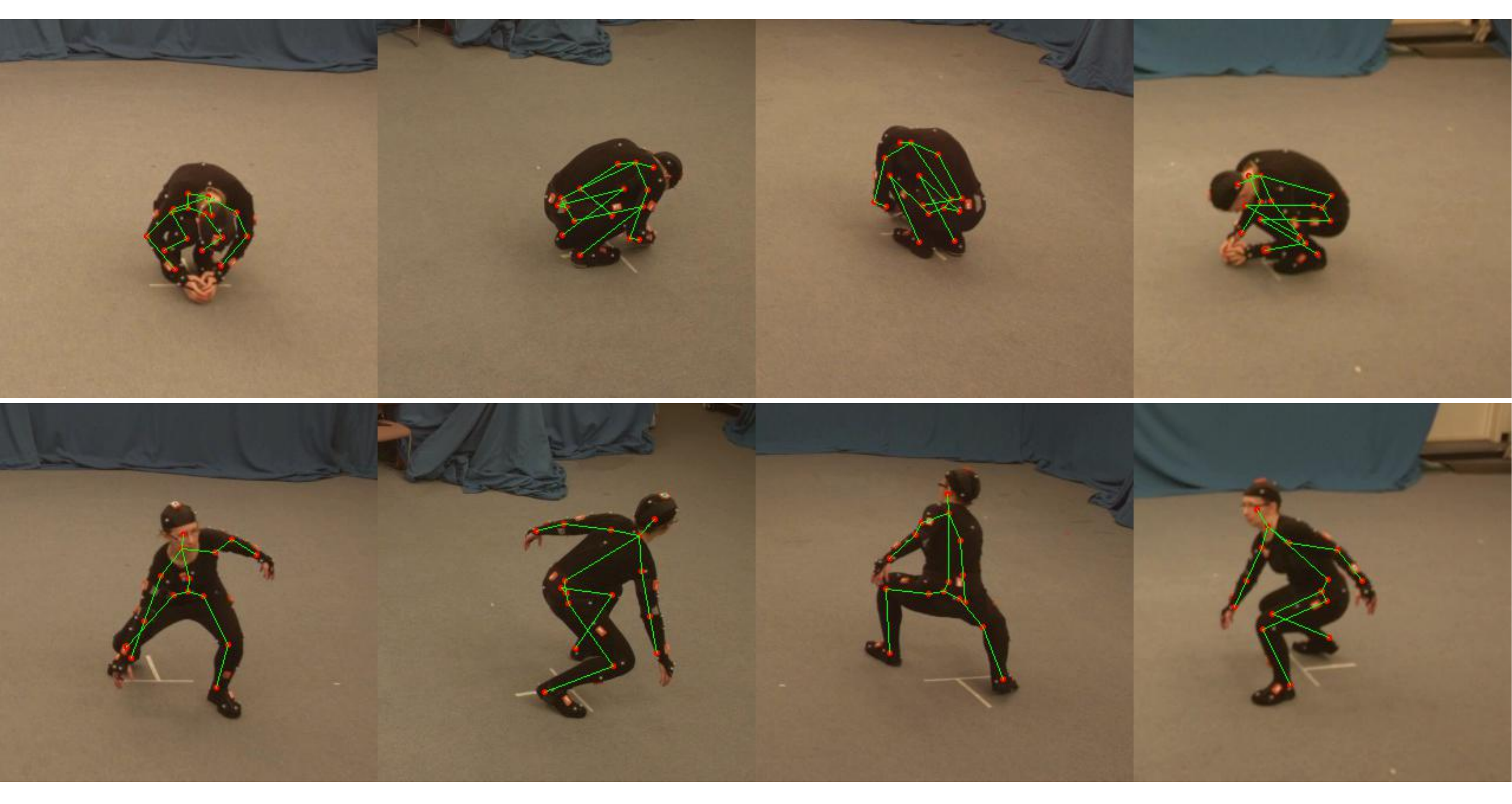}}
\caption{Self-occlusions introduce challenge in estimating 3D human pose from only vision data. IMUs can provide extra occlusion-free information. Here ground truth poses are shown for clarity.}
\label{fig_intro}
\end{center}
\end{figure}

3D human pose estimation is one of the most important problems in computer vision, which is closely related with human motion analysis, action recognition, human-computer interaction and so on \cite{navaratnam2005hierarchical,yang2017learning,gao2019dual-hand,arnab2019exploiting,liu2020feature}. The common solution of this problem is to predict the 3D keypoint coordinates of a predefined human skeleton from single-view \cite{martinez2017simple} or multi-view RGB images \cite{iskakov2019learnable,qiu2019cross}. In this solution, it is widely known that occlusions or self-occlusion in images introduce stochastic estimation error in extracting image features. As shown in Figure \ref{fig_intro}, the occlusion is inevitable in vision data. It may significantly reduce the performance and result in unreasonable human pose \cite{zhang2020object,kocabas2021pare}. 

To cope with this problem, recently an increasing number of approaches attempt to integrate extra data from other sensors, such as IMUs \cite{yi2021transpose,trumble2017total}. IMUs that are attached to human body can provide occlusion-free local information, which is valuable to improve the performance of 3D human pose estimation. Despite this, how to efficiently combine the data from vision and IMUs is intricate and challenging to explore. The main reason is that the information collected from cameras and IMU sensors is heterogeneous and hard to collaborate with each other in a fusion framework. 

Some previous methods partially alleviate the difficulty of IMU-vision fusion from two different aspects, but also have respective shortcomings. The first category of approaches \cite{zhang2020fusing,huang2020deepfuse} turn the different sensors data from heterogeneous into homogeneous. The IMU raw data is transformed to bone vectors \cite{zhang2020fusing} or volume features \cite{huang2020deepfuse} to more conveniently fuse with high-level deep image features under learning-based frameworks. Since the outputs of these frameworks only contain global 3D coordinates of human joints, the transformed IMU data fails to fuse kinematic information or help generate anatomically more reasonable skeletons. The second category of approaches \cite{gilbert2019fusing,von2016human} utilize the local rotation and acceleration data from IMUs to build energy terms and optimize generative models for pose estimation. Although these fusion methods employ IMU raw data, the generative models usually are not efficient enough compared to learning-based frameworks. Also, only low-level vision features such as silhouettes or estimated 2D poses are embedded into the optimization process.

To address the above-mentioned challenges, in this work we propose an IMU-vision sensor fusion framework for parametric 3D human pose estimation, named \emph{FusePose}. We first apply an algorithm called algebraic triangulation (AlgTri) \cite{iskakov2019learnable} to predict a sub-optimal 3D human pose from multi-view images (MVIs) as a baseline result, also as reliable high-level features for latter sensor fusion. Then, as performed in the previous work \cite{zhang2020fusing}, naively we calculate the bone vectors using IMU information and utilize them to help improve the baseline result in hard cases with self-occlusions. A threshold screening algorithm is proposed in this NaiveFuse method for filtering hard cases. Obviously, NaiveFuse neglects the value of IMU raw data. Thus, we propose the second method, i.e. Kinematic Fuse (KineFuse), to fuse IMU raw data with transformed baseline result in kinematic space. To be specific, we first build a parametric human skeleton model in kinematic space. Then we propose a canonical inverse kinematic (IK) layer in deep neural network to process the baseline 3D pose to pose parameters of the parametric human skeleton model. IMU raw data is calibrated and carefully aligned as the similar pose parameters which are finally fused with the pose parameters converted from baseline result. NaiveFuse and KineFuse are both test-time post-processing algorithms based on threshold screening. To further embed IMU information into network training process and increase the robustness of the whole framework, we propose the last approach, i.e. Adaptive Deep Fuse (AdaDeepFuse), which is end-to-end trained using IMU and vision data. This approach adaptively performs sensor fusion on not only hard cases but also other cases. Same as KineFuse, AdaDeepFuse outputs refined pose parameters, which are then fed into a forward kinematic (FK) layer to produce the final fused 3D pose. 

To sum up, IMU-vision fusion in kinematic space lies in the central place of KineFuse or AdaDeepFuse, where the data from heterogeneous modalities is aligned as homogeneous pose parameters, making the most advantage of complementary raw information for accurate and plausible parametric human pose results. In addition, there are two more merits of the proposed framework. The first one is that the consistency of the bone length can be ensured when fusing in kinematic space. This figures out an important issue in heatmap-based 3D pose estimation \cite{kamel2020hybrid,li2021tokenpose}, i.e. bone stretch due to varying illumination in images or quantization error in keypoint position generation. The second merit is that sensor fusion in kinematic space can aggregate twist rotation of bones in 3D human skeleton, which is incapable for sensor fusion in euclidean space.

Experiments demonstrate that our proposed methods achieve superior performance on Total Capture dataset which contains IMUs and multi-view images. Thanks to the well-designed algebraic triangulation algorithm, the baseline result already surpasses state-of-the-arts which only use vision information by a large margin. Compared to the baseline result, NaiveFuse, KineFuse and AdaDeepFuse further improve the pose estimation performance by $5.4\%$, $10.0\%$ and $13.1\%$, 
respectively. Also, we prove the generalization ability of the proposed framework on the new benchmark via experiments on Human3.6M dataset. In a nutshell, our contributions are summarized as follow:

\indent $\bullet$ We build a parametric human skeleton model which can perform interconversion of 3D pose and pose parameters via canonical inverse kinematic layer and forward kinematic layer in deep neural network, enabling fusion in kinematic space and end-to-end network training.

\indent $\bullet$ We propose three approaches, i.e. NaiveFuse, KineFuse and AdaDeepFuse, which effectively aggregate IMU raw data and reliable intermediate vision results, via leveraging the advantages of image clues in euclidean space and IMU rotations in kinematic space.

\indent $\bullet$ Through experiments, we demonstrate that: (1) the fusion in kinematic space simultaneously ensures global skeleton consistency and local rotation reasonability; (2) the proposed framework significantly improves 3D pose estimation performance under self-occlusion.

\section{Related Work}
 
\subsection{IMU-Vision Sensor Fusion}

Generally, IMU measures accelerations and angular velocities, then orientations can be solved leveraging filter algorithms \cite{bachmann2001inertial,foxlin1996inertial,del2018computationally,roetenberg2005compensation,vitali2020robust}. The commercial solution of \cite{xsens2009full} aggregates 17 IMUs to estimate human pose in a Kalman Filter. The seminal work of \cite{vlasic2007practical} uses a custom made suit to target human motion un everyday surroundings. There are also some works combine depth vision data with IMUs \cite{helten2013personalization,zheng2018hybridfusion}. The approach of \cite{malleson2017real} tries to integrate IMUs with 2D posees detected in one or two views. Reconstructing human poses from only IMUs is an under-constrained problem, as the information IMUs provide is insufficient for accurate human pose estimation. Thus, combining IMUs with vision data such as videos and images is worth employing. IMU-vision sensor fusion can take advantage of the supplementary strengths of the two data sources, i.e. the global drift-free position from vision data and local occlusion-free limb orientation even under fast motion from IMU data. The early work on the combination of video and inertial sensors is implemented by Pons-Moll \cite{pons2010multisensor} via optimizing a local hybrid tracker. Marcard \emph{et al.} \cite{von2016human} further develop this strategy under a setting of 8 viewpoint videos and 13 IMUs. Malleson \emph{et al.} \cite{malleson2019real} integrate constraints from IMUs, multi-view images and a human pose prior model to optimize generated parametric human pose. These methods explore the raw data of IMUs via minimizing the rotation or the acceleration energy terms. However, they only utilize rough vision cues such as image silhouettes or estimated 2D human pose.

As the deep networks ensure the high quality of extracted image features, deep feature-based methods for 3D human pose estimation develop well in recent years. The 2D heatmaps can serve as appropriate intermediate cues for either multi-view features aggregation or IMU-vision fusion. Gilbert \emph{et al.} \cite{gilbert2019fusing} back-project multiple viewpoint features to probabilistic visual hull (PVH). The IMU orientations are processed to 3D joint locations by a forward kinematic solver and then embedded into 3D pose estimation via tensor concatenation. Huang \emph{et al.} \cite{huang2020deepfuse} propose to transform IMU-bone orientations to volume features, which are then concatenated with vision volume heatmaps. Zhang \emph{et al.} \cite{zhang2020fusing} process the IMU orientations to bone vectors for either promoting the 2D joint heatmap refinement in terms of multi-view geometry or optimizing the 3D occupancy of joints via picture structure model (PSM). Although high accuracy the above-mentioned deep learning-based methods achieve, they do not leverage the raw orientation and acceleration data from IMUs. Instead, they utilize the transformed IMU information such as bone vectors for the convenience of fusing with deep vision features. 

In this work, we fuse the IMU raw data and reliable intermediate features of multi-view vision data under a parametric human pose representation, ensuring both the efficiency of sensor fusion and the kinematic consistency of estimated 3D human skeletons.
 
\subsection{Parametric Human Pose Estimation}

3D human pose estimation is often formulated as learning from sensor data such as monocular or multi-view images to estimate the 3D keypoint coordinates of a human skeleton \cite{li2021human,iskakov2019learnable,martinez2017simple}. This location-based strategy recognizes each human joint as a point from heatmap and each bone as a line from a predefined human kinematic tree. However, there exist problems unsolved under this point-line human pose representation. First, directly regressing human joints from heatmaps which are generated by convolutional neural network may introduce bone stretching issue\cite{holden2016deep,holden2017phase} or left-right asymmetry\cite{dabral2018learning}. Second, the twist rotation of the bone is ignored and can not be integrated into the whole framework, resulting in that the estimated 3D human pose may be inadequate to represent the true human motion\cite{li2021hybrik}. Thus, a parametric human skeleton is paramount for more plausible human pose.

Parametric human pose\cite{pavlakos2019expressive,grochow2004style,liu2013markerless,vlasic2008articulated} represents the human body with skeletons with a certain number of keypoints and the parameters attached to them. The pose parameters are often based on a human model with local coordinates for joints or bones. This model-based methods have great merit for more than human skeleton estimation such as robotic control\cite{csiszar2017solving}, motion retargeting\cite{villegas2018neural}, and human mesh recovery\cite{kanazawa2018end,bogo2016keep,fan2021revitalizing,mehta2017vnect}. The works on model-based parametric human pose estimation can be divided into two categories: optimization-based methods and direct regression methods.

Optimization-based methods mainly focus on searching for the optimal solution of human pose parameters\cite{ye2014real}. The approach of \cite{lassner2017unite} fit SMPL \cite{loper2015smpl} to 2D detections via solving optimization problem. Energy functions are built to minimize the difference between generated human model and image features. The solution procedure often needs good initialization and a relatively long time for iteratively optimization, while mismatching the requirement of real scene application. 

With the development of deep networks, increasing studies focus on directly regressing either location-based pose\cite{iskakov2019learnable,martinez2017simple,sun2018integral} or parametric pose\cite{villegas2018neural,shi2020motionet,pavllo2019modeling,kanazawa2018end} in an end-to-end manner. Kanazawa \emph{et al.} \cite{kanazawa2018end} propose Human Mesh Recovery (HMR) to reconstruct a mesh representation that is parameterized shape and 3D joint angles. Shi \emph{et al.} \cite{shi2020motionet} proposed an approach that learns to predict joint rotations directly from training data. Despite the great contribution made by \cite{kanazawa2018end} and \cite{shi2020motionet}, these methods rely heavily on an extra adversarial loss to learn real-or-fake prior information and ensure that the regressed results lie on the manifold of natural pose parameters.


\section{Method}
\label{sec:method}
\subsection{Formulation}
\label{subsec:pre}

\noindent \textbf{Parametric pose representation.}
\quad A 3D human skeleton is constituted of 3D locations of $J$ keypoints $P = \lbrace p_{j} \rbrace_{j=1}^{J}$ and the certain connections between them. The parametric pose representation formulates a rigid human body as a rest T-pose skeleton $P^{T}$ together with the root transition parameter $T_{1}$ and the rotation parameters $\boldsymbol{R} = \lbrace R_{j} \rbrace_{j=2}^{J}$ of the local systems $S = \lbrace s_{j} \rbrace_{j=2}^{J}$ attached to the bone vectors. 

From the rest skeleton $P^{T}$, we can initialize all the local systems $S$ by defining cartesian coordinates. A vector $b^{g}$ in global system can be transferred to $s_j$ by $b^{j} = R_{j}^{g} b^{g}$, where $R_{j}^{g}$ denotes the rotation from global system $s_{g}$ to local system $s_{j}$. Also, the relative rotation from the parent $pa(j)$ of $j$ to $j$ can be calculated by:
\begin{equation}
R_{j}^{pa(j)} = R_{j}^{g} (R_{pa(j)}^g)^{-1} \label{equa1}.
\end{equation}

\begin{figure}[t]
 \begin{center}
 	\centerline{\includegraphics[width= 7cm ]{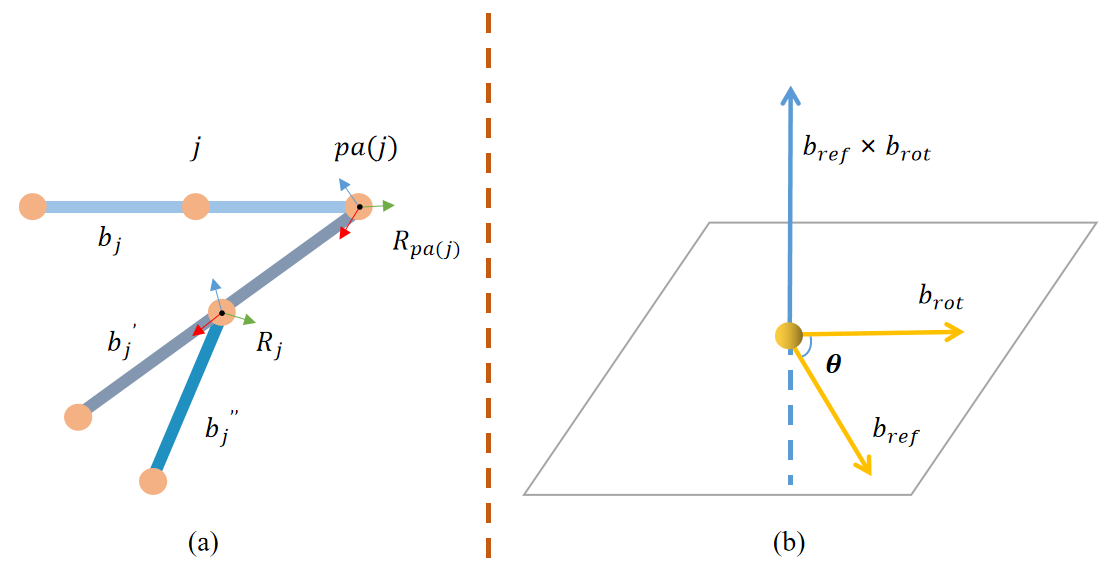}}
\caption{(a) In the process of FK, $b_{j}$ is first rotated to $b_{j}^{'}$ under the rotation $R_{j}$ of joint $j$, then is further rotated to $b_{j}^{''}$ under the rotation $R_{pa(j)}$ of joint $pa(j)$. (b) Illustration of canonical solution of IK. The rotation axis is set as the cross product result of reference and rotated bone vector. The rotation angle is the angle between the two bone vectors.}
\label{fig2}
\end{center}
\end{figure}

~\\
\noindent \textbf{FK and IK.}
\quad Given the parameters $T_{1}$ and $\boldsymbol{R}$ , 3D human pose can be rebuilt by forward kinematics (FK). This process ensures the bone length consistency and simultaneously is able to integrate the twist rotation of the bone vector.

The purpose of FK is to reconstruct 3D joint locations from pose parameters along the kinematic chain. For example, all the joints in the path from joint $j$ to root joint are formulated as $\lbrace pa^{i}(j) \rbrace_{i=1}^{n_{j}}$, then the bone vector in global system $b_{j}^{g}$ can be calculated by:
\begin{equation}
b_{j}^{g} =  
R_{pa^{n}(j)}^{g} 
R_{pa^{n}(j)} 
R_{pa^{n-1}(j)}^{pa^{n}(j)}
...
R_{pa(j)}
R_{j}^{pa(j)}
R_{j} 
b_{j}^{j} \label{equa2},
\end{equation}
where the bone vector in local system $b_{j}^{j}$ and all the relative rotation are calculated from the rest T-pose, as described above. $n_{j}$ is simplified as $n$ here and later. The 3D location of $j$ therefore can be calculated by:
\begin{equation}
p_{j} = p_{pa(j)} + b_{j}^{g} \label{equa3}.
\end{equation}
As illustrated in Figure \ref{fig2} (a),  the bone vector $b_{j}$ from $pa(j)$ to $j$ is not only affected by the rotation $R_{j}$ of its related joint $j$, but also by the rotations of all the upstream joints $\lbrace pa^{i}(j) \rbrace_{i=1}^{n}$, matching the kinematics of rigid bodies and the real anatomic motion of human. 

While the forward kinematics is well-defined and easy to solve, the inverse kinematics (IK) is an ill-posed problem because the 3D locations of joints fail to model the rotation information. In this work, we propose a canonical solution of IK based on axis-angle representation.

\begin{figure*}[ht]
 \begin{center}
 	\centerline{\includegraphics[width= \linewidth]{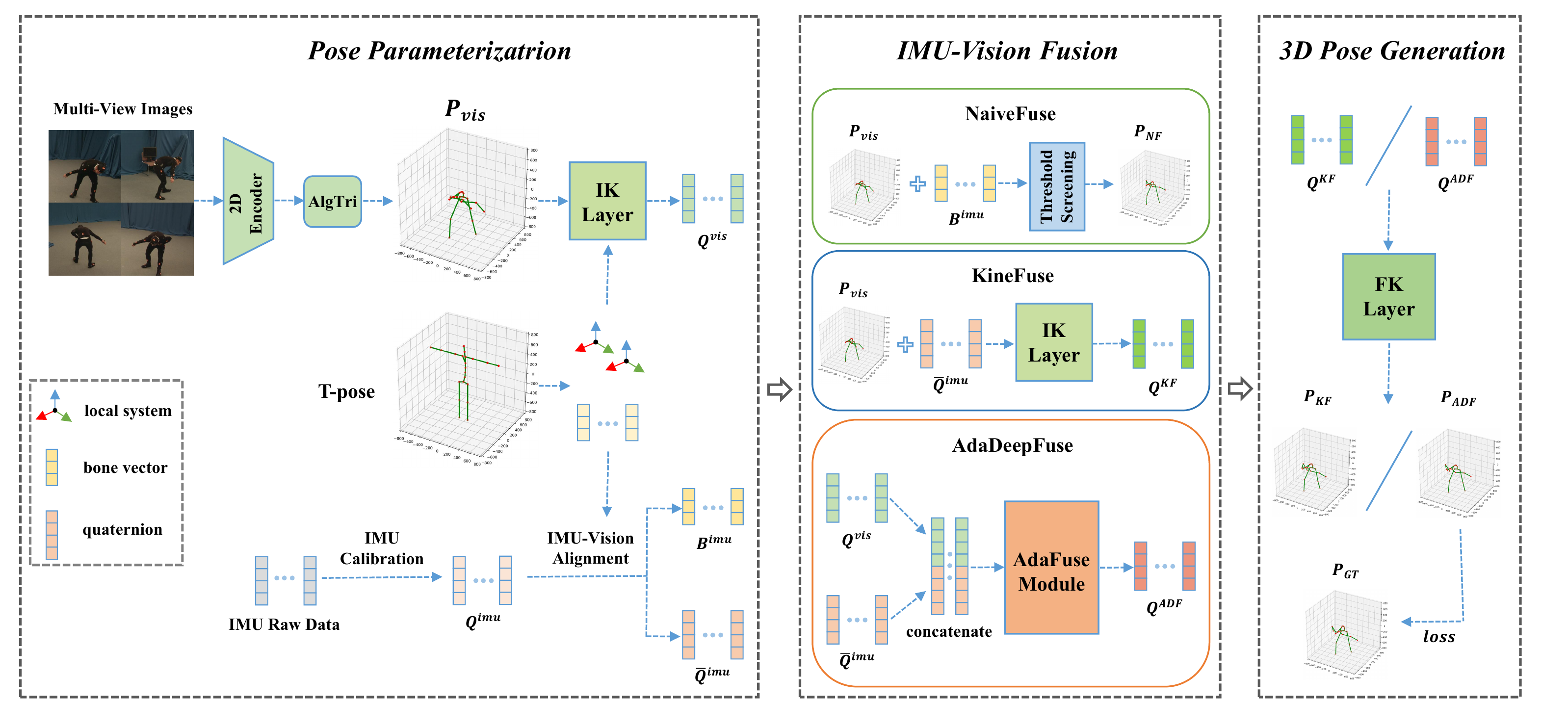}}
\caption{The proposed IMU-vision fusion framework for parametric 3D human pose estimation. The 3D pose $P^{vis}$ estimated from multi-view images is parameterized to local rotations $Q^{vis}$ via IK layer. The IMU raw data is also processed to local rotations $\bar{Q}^{imu}$ or bone vectors $B^{imu}$ and is integrated with vision data ($P^{vis}$ or $Q^{vis}$) by three fusion approaches: \emph{NaiveFuse}, \emph{KineFuse} and \emph{AdaDeepFuse}. The final fused 3D poses of \emph{KineFuse} and \emph{AdaDeepFuse} are generated via FK layer. Only \emph{AdaDeepFuse} needs to be trained end-to-end.}
\label{fig:framework}
\end{center}
\end{figure*}

Same as the process of FK, Inverse Kinematics (IK) is performed along the kinematic chain. When calculating the rotation parameter $R_{j}$ of joint $j$, the rotation parameters of all his parents $\lbrace pa^{i}(j) \rbrace_{i=1}^{n}$ are supposed be already solved. As the bone vector $b_{j}^{g}$ can be derived from the current 3D pose, the only unknown variable in Eq. (\ref{equa2}) is $R_{j}$. Figure \ref{fig2} (b) shows how we solve a canonical axis-angle representation of the local rotation from two 3D bone vectors. The axis $K$ and angle $\theta$ are then transferred to the rotation matrix by the Rodriguez formula: 
\begin{equation}
R = I + sin(\theta)K + (1-cos(\theta))K^{2}.
\end{equation}
Despite this solution is unable to represent the true rotation of the local system related to the bone vector, it provides a unique mapping from 3D joints location to pose parameters. A more detailed introduction of IK will be expanded in section \ref{subsec:fusion} and Alg \ref{alg1}.

Since the IK and the FK are both differentiable, we implement them as two parameter-free layers in the deep neural network: IK Layer and FK Layer. Thus, they are both lightweight enough to be inserted into any 3D pose estimation framework as a useful module.

The whole framework of our approach is illustrated in Figure \ref{fig:framework}. We first process the information from whether multi-view images (MVIs) or IMUs to pose parameters based on the kinematic human model. Then, a total of three methods are proposed to fuse those pose parameters for bolstering the performance of 3D human pose estimation. In the rest of this section, we will elaborate on pose parameterization and the three parts of the whole frameworks, relatively.

\subsection{Pose Parameterization}

~\\
\noindent \textbf{Multi-view images.}
\quad We adopt an effective architecture to predict a sub-optimal 3D human pose $P^{vis}$ from MVIs. Following \cite{iskakov2019learnable}, we use ResNet as our 2D encoder to generate 2D joint heatmaps, and a differential algebraic triangulation (AlgTri in Figure \ref{fig:framework}) with confidence weights of different views to obtain a linear approximate solution of 3D joint locations via Singular Value Decomposition (SVD). 

After the $P^{vis}$ is obtained, pose parameters are generated by an IK layer. The root global transition $T_{1}$ is determined as the root location in $P_{vis}$. For better properties of computation consistency and stability, following \cite{pavllo2020modeling}, we convert the rotation matrix of local system $R_{j}^{vis}$ to quaterion $q_{j}^{vis}$ as pose representation parameters.

The pose parameters $Q^{vis} = \lbrace q_{j}^{vis}\rbrace_{j=2}^{J}$ have certain distributions based on the articulation range of human joints with known degree of freedoms (DoFs). The main reason we explore the pose parameters in this work instead of 3D joint locations is to more conveniently and efficiently aggregate the pose information from MVIs and IMUs. We will elaborate on the detail of the aggregation algorithms in section \ref{subsec:fusion}.

~\\
\noindent \textbf{IMUs.}
\quad IMUs are rigid sensors attached to human bones, measuring the local rotation by integrating the recorded gyroscope data from the reference frame to the current frame. The information from IMUs is often represented as quaternion. In order to employ the global rotation $q_{k}^{imu}$ of human bone attached with IMU $k$, the raw data $q_{k}$ from the IMU need to be calibrated using the IMU-bone offset $q_{kb}$ and the IMU reference frame-global offset $q_{kg}$ by:
\begin{equation}
q_{k}^{imu} = (q_{kb})^{-1} \otimes q_{kg} \otimes q_{k},
\end{equation}
where  $\otimes$ denotes the quaterion multiplication.

The calculated bone global rotation $Q^{imu} = \lbrace q_{k}^{imu} \rbrace_{k=1}^{K}$ is still difficult to collaborate with the pose parameters generated from MVIs. Hence, it is also indispensable to align the IMU and vision information as the same representation for robust and efficient consolidation. There exist two available operations for IMU-vision alignment. The first one is directly applying the global rotations from IMUs to the bone vectors in T-pose, and generating the current frame bone vectors $B^{imu} = \lbrace b_{k}^{imu} \rbrace_{k=1}^{K}$ which are the same representation with the bone vectors derived from 3D joint locations. $K$ is the number of IMUs attached to human bones. In this work, $K$ is set as 8 following \cite{zhang2020fusing}, employing IMUs in thighs, calves, upper arms and forearms. The bone vector rotation is performed by $b^{imu}_{k} = \bar{q}_{k}^{imu} \rhd b_{j}^{j}$, where $\rhd$ denotes applying the rotation of a quaternion on a 3D vector and $j$ is the endpoint of the bone $k$ attached with IMU. The second operation for IMU-vision alignment is converting the global rotation $Q^{imu}$ to the local rotation $\bar{Q}^{imu} = \lbrace \bar{q}_{k}^{imu} \rbrace_{k=1}^{K}$ by:
\begin{equation}
\bar{q}_{k}^{imu} = q_{k}^{imu} \otimes (q_{j}^{g})^{-1},
\end{equation}
where $q_{j}^{g}$ denotes the transformation from global $s_{g}$ to local $s_j$. Note that the local rotations are also part of the unknown variables in the IK process. Thus, $\bar{Q}^{imu}$ can be efficiently embedded into the IK layer and can collaborate well with pose parameters $Q^{vis}$ from vision data.

\subsection{IMU-Vision Fusion Framework}
\label{subsec:fusion}

As illustrated in the middle of Figure \ref{fig:framework}, we propose three approaches to fusing information from vision and IMUs. Next, we will dive into the elaboration of these three approaches.

~\\
\noindent \textbf{NaiveFuse.}
\quad The first approach NaiveFuse aggregates the estimated coarse 3D pose $P^{vis}$ from multi-view images and the rotated bone vectors $B^{imu}$ using IMU raw data. There is a large consent that the IMU data suffers from drift error due to the integral operation in data collection. Directly replacing the bone vectors in $P^{vis}$ using $B^{imu}$ may result in pose error increasing for easy-to-estimate frames. Thus, we propose a threshold screening method based on vectorial angle to decide if the bone vector should be replaced or not as follows:
\begin{equation}
b^{vis}_{j} = \begin{cases}
   b^{imu}_{k}, &\text{if } \theta_{k} > \theta_{t} \\
   b^{vis}_{j}, &\text{otherwise},
\end{cases}
\end{equation}
where $\theta_{k}$ is the angle between $b^{imu}_{k}$ and  $b_{j}^{vis}$ and $\theta_{t}$ is the threshold.

Obviously, this approach can only facilitate the results of joints related to IMUs by a decent margin. Assuming there exist large estimation errors on the upstream joints, NaiveFuse can hardly improve the results. Another drawback of NaiveFuse is that it is inadequate to just naively utilize the transformed 3D bone vector to reconstruct the real motion parameter of human bone which is a rigid body. For better leveraging the rigid bone rotation information from IMUs on a deeper level, we propose another fusion method in a kinematic manner.

\begin{algorithm}[t] 
\caption{Kinematic Fuse}
\begin{algorithmic}
\STATE \textbf{Input:} $P^{vis}, \bar{Q}^{imu}, P^{T}$
\STATE \textbf{Output:} $Q^{KF}, P^{KF}$
\STATE \hspace{0.5cm} $T_{1} \gets p_{1}^{vis}$ in $P^{vis}$ 
\STATE \hspace{0.5cm} $p_{1}^{KF} \gets T_{1}$
\STATE \hspace{0.5cm} \textbf{for} $j$ along the kinematic tree \textbf{do}
\STATE \hspace{1cm} $b_{j}^{vis} \gets (p_{j}^{vis} - p_{pa(j)}^{vis})$
\STATE \hspace{1cm} $q_{j}^{total} \gets$ CanonicalSolve($b_{j}^{vis}, b_{j}^{j}$)
\STATE \hspace{1cm} \textbf{if} $b_{j}$ is attached with IMU $k$ \textbf{then}
\STATE \hspace{1.5cm} $b_{k}^{imu} \gets \bar{q}_{k}^{imu} \rhd b_{j}^{j}$
\STATE \hspace{1.5cm} $\theta_{k} \gets$  $<b_{k}^{imu}, b_{j}^{j}>$
\STATE \hspace{1.5cm} \textbf{if} $\theta_{k} > \theta_{t}$ \textbf{then}
\STATE \hspace{2.0cm} $q_{j}^{total} \gets \bar{q}_{k}^{imu}$
\STATE \hspace{1.5cm} \textbf{end if}
\STATE \hspace{1cm} \textbf{end if}
\STATE \hspace{1cm} $q_{j} \gets (q_{pa^{n}(j)}^{g})^{-1} \otimes q_{j}^{total}$
\STATE \hspace{1cm} \textbf{for} $pa^{i}(j): i \in [n,1] $ \textbf{do}
\STATE \hspace{1.5cm} $q_{j} \gets (q_{pa^{i}(j)})^{-1} \otimes q_{j}$
\STATE \hspace{1.5cm} $q_{j} \gets (q_{pa^{i-1}(j)}^{pa^{i}(j)})^{-1} \otimes q_{j}$
\STATE \hspace{1cm} \textbf{end for}
\STATE \hspace{0.5cm} \textbf{end for}
\STATE \hspace{0.5cm} $Q^{KF} \gets \lbrace q_{j} \rbrace_{j=2}^{J}$
\STATE \hspace{0.5cm} $P^{KF} \gets FK(Q^{KF}, P^{T})$
\end{algorithmic}
\label{alg1}
\end{algorithm}

~\\
\noindent \textbf{KineFuse.}
\quad Kinematic Fuse (KineFuse) aggregates the pose parameters $Q^{vis}$ calculated from $P^{vis}$ via IK layer and the aligned IMU rotation information $\bar{Q}^{imu}$. The main advantage of utilizing local rotations in KineFuse instead of bone vectors in NaiveFuse is that the twist rotations of bones can be compensated in the IK process. As described in section \ref{subsec:pre}, the proposed canonical solution of IK is ill-posed and unable to disentangle the twist rotation from two 3D bone vectors. Despite that the twist rotation of a bone has no impact on the 3D location of joints attached to the bone, the situation is different for the lower downstream joints. For instance, the twist rotation of thighs is unable to affect the location of knees while able to affect that of ankles. 

The whole process of KineFuse is summarized in Alg \ref{alg1}. Given the predicted 3D pose $P^{vis}$ from vision, the calibrated and aligned IMU rotation information $\bar{Q}^{imu}$ and the rest T-pose $P^{T}$, the KineFuse is performed based on IK layer and output the fused pose parameters $T_{1}$ and $Q^{KF}$. In order to get to the final fused 3D pose $P^{KF}$, the fused pose parameters are fed into FK layer as described in section \ref{subsec:pre}.

The most important improvement in KineFuse is that it can simultaneously ensure the consistency of bone length and the accuracy of limb joints estimation. To be specific, the results of joints that are not related to IMUs can also be improved on the ground that the bone stretching errors are eliminated via IK and FK process. Furthermore, the estimation errors of joints that are related to IMUs can be explicitly reduced on the basis of more accurate ancestor joints, together with IMU information which entangles with twist rotation.

However, in KineFuse there exists an important issue that, it just serves as a post-processing approach in test-time, same as NaiveFuse. IMU information should also be utilized in the network training stage to help increase the robustness of the whole framework. Thus, in the last approach AdaDeepFuse, we explore efficient IMU-vision fusion by a deep learning-based supervised way.

~\\
\noindent \textbf{AdaDeepFuse.} Adaptive Deep Fuse (AdaDeepFuse) receives pose parameters from MVIs $Q^{vis}$ and IMUs $\bar{Q}^{imu}$, which are then concatenated and fed into an AdaFuse module. The AdaFuse module is implemented as a multi-layer perceptron (MLP) with the output of fused pose parameters $Q^{ADF}$. Then, same as in KineFuse $Q^{ADF}$ will be utilized to infer an IK layer for the final fused 3D pose $P^{ADF}$.

Different from NaiveFuse and KinFuse, AdaDeepFuse are trained in an end-to-end manner with supervision in either the 3D human pose or pose parameters. We claim that the error from IMUs and 3D pose estimated from MVIs can be adaptively compensated via training a neural network. The AdaFuse module can determine what extent $Q^{vis}$ or $\bar{Q}^{imu}$ contributes to more accurate results. We fuse pose parameters instead of bone vectors since the quaternion space is continuous for interpolation and robust for network training. The total loss $L$ function for training AdaDeepFuse consists of two terms of loss function for 3D human pose $L_{pose}$ and loss function for pose parameters $L_{param}$ as :
\begin{equation}
L = L_{pose} + \alpha L_{params},
\end{equation}
where $\alpha$ is the weight of pose parameter loss and set as $1 \times 10^{-2}$ in this work. Both the two loss terms are implemented as computing the Smoothed L1 loss of results compared to ground truths. The ground truth 3D pose $P^{GT}$ is collected from MoCap equipment. The ground truth pose parameters $Q^{GT}$ are derived from $P^{GT}$ via IK layer.
\section{Experiments}

In this section, we conduct the experiments to evaluate the proposed IMU-vision fusion framework. First, the two datasets used in the experiments are introduced. Then, the comparisons between our methods with other methods are reported. Finally, the ablation studies are conducted.

\subsection{Datasets and Experiment Settings}

\noindent \textbf{Total Capture} \cite{trumble2017total}.
\quad This dataset is a large-scale benchmark containing information of 13 IMUs, multi-view videos and 3D human pose ground truth. Following \cite{zhang2020fusing,trumble2017total}, we use four (1, 3, 5, 7 ) of whole eight views in this work for efficiency. For the number of chosen IMUs, we design two optional settings: 8 IMUs on the limbs and 4 of them on either the upper limbs or lower limbs. We partition the training and testing dataset with respect to subjects and performance sequence. The training set consists of performances ROM1, 2, 3; Walking1, 3; Freestyle1, 2; Acting1, 2 and Running1 on subjects 1, 2 and 3. The testing set contains the performances Freestyle3, Acting3 and Walking2 on all subjects. We use all frames in the training set and every eighth frame in the testing set. We evaluate all the three proposed approaches on Total Capture dataset. To be specific, we supervise NaiveFuse and KineFuse only on the predicted 3D pose $P^{vis}$ from MVIs, and supervise AdaDeepFuse on the fused 3D pose $P^{ADF}$.

~\\
\noindent \textbf{Human3.6M} \cite{ionescu2013human3}.
\quad We use this dataset for generalization evaluation. It consists of 3.6M frames from 4 synchronized cameras with the 3D pose ground truth. There are 11 subjects in Human3.6M. Following previous work \cite{sun2018integral}, we split it into the training set (S1, S5, S6, S7, S8) and testing set (S9, S11). Since it has no IMU sensors, we calculate bone vectors from 3D pose annotations as the simulative IMU information. Unlike real IMU sensor data, this kind of data is absolutely accurate, free from drift or other errors and unsuitable for training a model. Thus, we just verify NaiveFuse and KineFuse in Human3.6M dataset.

~\\
\noindent \textbf{Implementation details.}
\quad We use ResNet-152 \cite{he2016deep} as 2D encoder backbone, initialized with ImageNet pre-trained weights. In Total Capture we train the 2D encoder and the AdaFuse Module end-to-end from scratch for 15 epochs using Adam\cite{kingma2014adam} optimizer, while in Human3.6M we utilize the pre-trained AlgTri model weights in \cite{iskakov2019learnable}. The input multi-view images are resized to $320 \times 320$ in Total Capture and $384 \times 384$ in Human3.6M. We calculate the Mean Per Joint Position Error (MPJPE) as the metric to evaluate the performance of 3D pose estimation. It measures the mean distance between estimated 3D joint locations and ground truths over all subjects and frames of the testing set.

\begin{table*}[!tp]
\caption{Comparison of the 3D pose estimation errors MPJPE (mm) of different methods on the Total Capture dataset. Our method outperforms previous state-of-the-arts.}
\begin{center}
\resizebox{\textwidth}{!}{
\begin{tabular}{lccccccccc}
\toprule
Method & train w/ IMUs & test w/ IMUs &\multicolumn{3}{c}{SeenSubject (S1,2,3)} &\multicolumn{3}{c}{UnseenSubject (S4,5)} & Average  \\
 &&&W2&A3&FS3&W2&A3&FS3&\\
\midrule
Tri-CPM \cite{yang2017learning} & & &79.0 &112.0&106.0 &79.0 &149.0 &73.0 &99.0\\
PVH \cite{trumble2017total} & & &48.3 &94.3&122.3 &84.3 &154.5 &168.5 &107.3\\
LSTM-AE \cite{trumble2018deep} & & &13.0 &23.0&47.0 &21.8 &40.9 &68.5 &34.1\\
IMUPVH \cite{gilbert2019fusing} &\checkmark&\checkmark&19.2 &42.3&48.8 &24.7 &58.8 &61.8 &42.6\\
Fusion-RPSM \cite{qiu2019cross} & & &19.0 &21.0&28.0 &32.0 &33.0 &54.0 &29.0\\
DeepFuse-Vision\cite{huang2020deepfuse} & & &- &-&- &- &- &- &32.7\\
DeepFuse-IMU \cite{huang2020deepfuse} &\checkmark& &- &-&- &- &- &- &28.9\\
GeoFuse-SN-ORSPM \cite{zhang2020fusing} & &\checkmark&- &- &- &- &- &- &25.5\\
GeoFuse-ORN-ORSPM \cite{zhang2020fusing} &\checkmark&\checkmark&14.3 &17.5 &\textbf{25.9} &23.9 &27.8 &49.3 &24.6\\
\midrule
AlgTri (baseline) & & &9.6&15.3&30.9&24.1&30.6&61.3&25.9\\
NaiveFuse (ours) & &\checkmark &9.6&14.8&27.9&23.9&30.4&55.4&24.5\\
KineFuse (ours) & &\checkmark &\textbf{9.5}&14.3&26.8&22.4&28.7&51.8&23.3\\
AdaDeepFuse (ours) & \checkmark& \checkmark &10.2&\textbf{13.7}&\textbf{26.3}&\textbf{21.7}&\textbf{26.8}&\textbf{49.2}&\textbf{22.5}\\
\bottomrule
\end{tabular}}
\end{center}
\label{table5}
\end{table*}

\subsection{Comparison to Other Methods}

We first compare the 3D human pose estimation performance of our framework to the state-of-the-arts on the Total Capture dataset. The results are listed in Table \ref{table5}. The last four rows show the methods implemented in this work, including the baseline AlgTri algorithm and three proposed IMU-vision sensor fusion approaches. Among all previous methods, first we can see that the \emph{LSTM-AE}\cite{trumble2018deep} and \emph{IMUPVH}\cite{gilbert2019fusing} achieve decent performance in the case of using temporal information. The error the state-of-the-art among the methods that only utilize vision data and single frame as input, i.e. \emph{Fusion-RPSM}\cite{qiu2019cross}, is $29mm$ and larger than $25.9mm$ achieved by our trained baseline AlgTri. This is because the multiple viewpoints information is aggregated well using view confidences weighted triangulation algorithm. The state-of-the-art performance using only vision data is boosted by 14.3\%. 

Among the methods that perform IMU-vision sensor fusion, our approach NaiveFuse ($24.5mm$), KineFuse ($23.3mm$) and AdaDeepFuse ($22.5mm$) achieve state-of-the-art performances. There are three main points need to be noted. First, the NaiveFuse and KineFuse both utilize IMU information for only testing and serve as post-processing approaches. They can be conviniently applied to any existing 3D human pose estimation framework. They both outperforme the method of \emph{GeoFuse-SN-ORPSM}\cite{zhang2020fusing} ($25.5mm$) which also only uses IMU for testing only. Second, the last proposed approach AdaDeepFuse uses IMU for both training and testing. It outperforms the method of \emph{GeoFuse-ORN-OPRSM}\cite{zhang2020fusing} ($24.6mm$). We increase the performance by 8.5\%. Third, the Recursive Pictorial Structure Model (RPSM) in \cite{zhang2020fusing} is proposed to perform 3D IMU-vision information integration based on enumeration algorithm, which is time-consuming and relies heavily on computing and memory resources. However, our approaches are computationally friendly since the proposed IK and FK layers are both lightweight and parameter-free. The AdaFuse module is also a shallow network. Thus, our framework has a wider spectrum of applications in the real scene.

\subsection{Ablation Study}

\begin{figure}[!tp]
 \begin{center}
 	\centerline{\includegraphics[width= \linewidth]{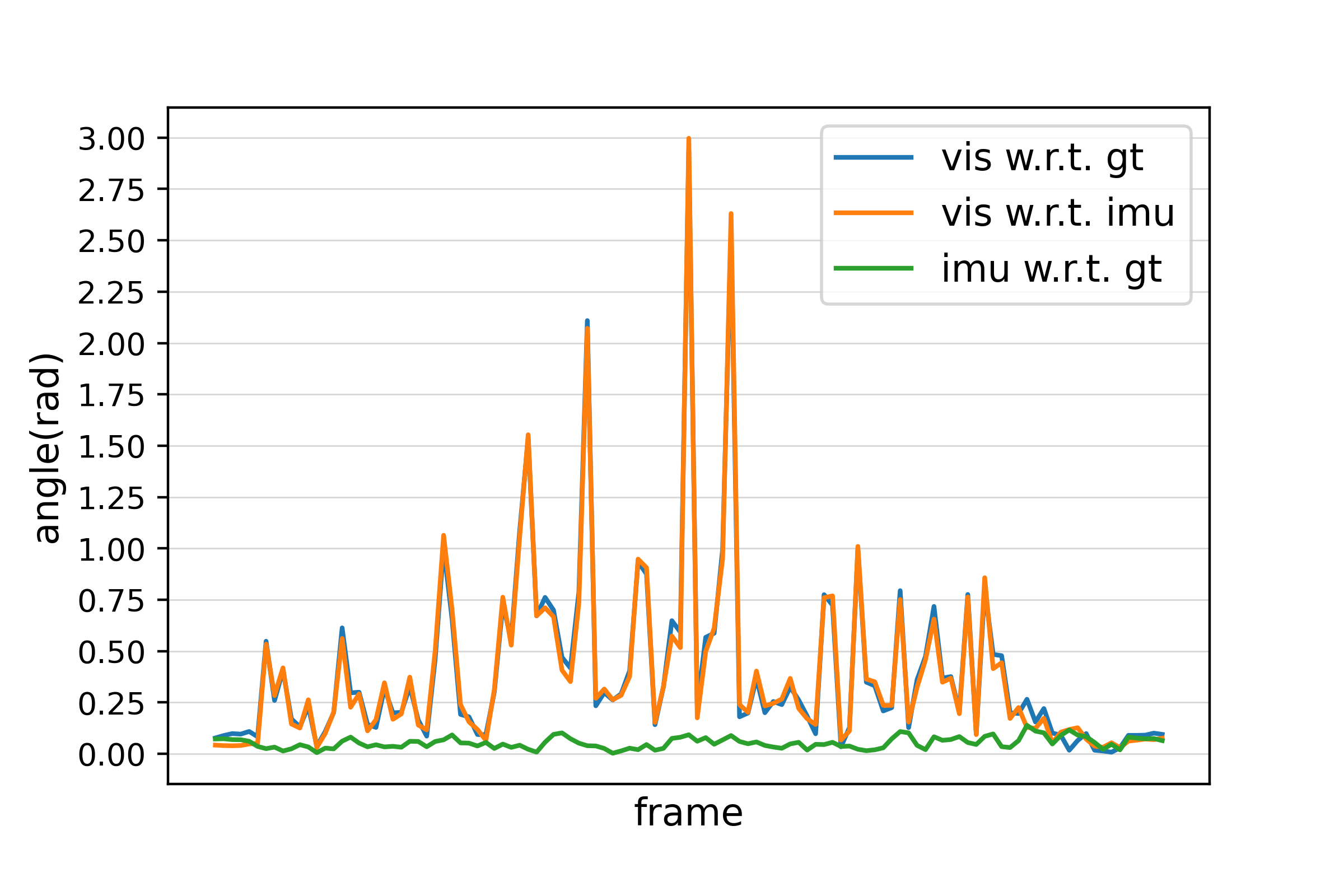}}
\caption{The vectorial angles between bone vectors which are derived from vision information, IMUs and ground truth. The curves show the angles of right forearm along a motion sequence.}
\label{fig4}
\end{center}
\end{figure}

\noindent \textbf{Threshold screening.}
\quad One of the paramount operations in NaiveFuse and KineFuse is the threshold screening for hard-to-estimate frames. To verify the necessity and robustness of this operation, we calculate vectorial angles between the bone vectors which are derived from vision, IMUs and ground truth. Note that the information from vision here is the estimated 3D pose $P^{vis}$ by AlgTri. Figure \ref{fig4} illustrates the angles curves of the right forearm bone vector in one of the motion sequences. By observing the curves, we can draw a conclusion that the difference between IMU and vision information is very close to the difference between IMU and ground truth. Hence, we can use a threshold to screen the proper hard cases in terms of the vectorial angle between bone vectors derived from IMU and vision. In addition, it can also be observed that the difference between IMU and ground truth is tiny, indicating the reliability of using IMU in test-time.

As shown in Figure \ref{fig4}, the angle curve between vision and IMUs is undulate due to pose estimation errors of $P^{vis}$. An appropriate value of threshold $\theta^{t}$ is crucial for figuring out suitable hard cases and then integrating IMU information to improve the performance. We randomly select $30 \%$ frames from Total Capture training set to perform an ablation study to determine the value of $\theta^{t}$. We list some possible values of $\theta^{t}$ and their respective estimation errors in Table \ref{table1}. We find that $\theta^{t} = 0.25$ achieves the best performance in our experiment setting. Note that NaiveFuse and KineFuse are employed in this experiment.

\begin{table}[!tp]
\caption{Different value of $\theta^{t}$ for methods that apply threshold screening and their respective 3D pose estimation errors (mm) in random selected validation set.}
\begin{center}
\scalebox{0.48}{
\resizebox{\textwidth}{!}{
\begin{tabular}{cccccc}
\toprule
$\theta^{t}$ & 0.15 & 0.20 & 0.25 & 0.30 & 0.35 \\
\midrule
NaiveFuse &24.83&24.61&\textbf{24.51}&24.64&24.79\\
KineFuse &23.58&23.38&\textbf{23.31}&23.37&23.51\\
\bottomrule
\end{tabular}}}
\end{center}
\label{table1}
\end{table}

\begin{figure}[!tp]
 \begin{center}
 	\centerline{\includegraphics[width= \linewidth]{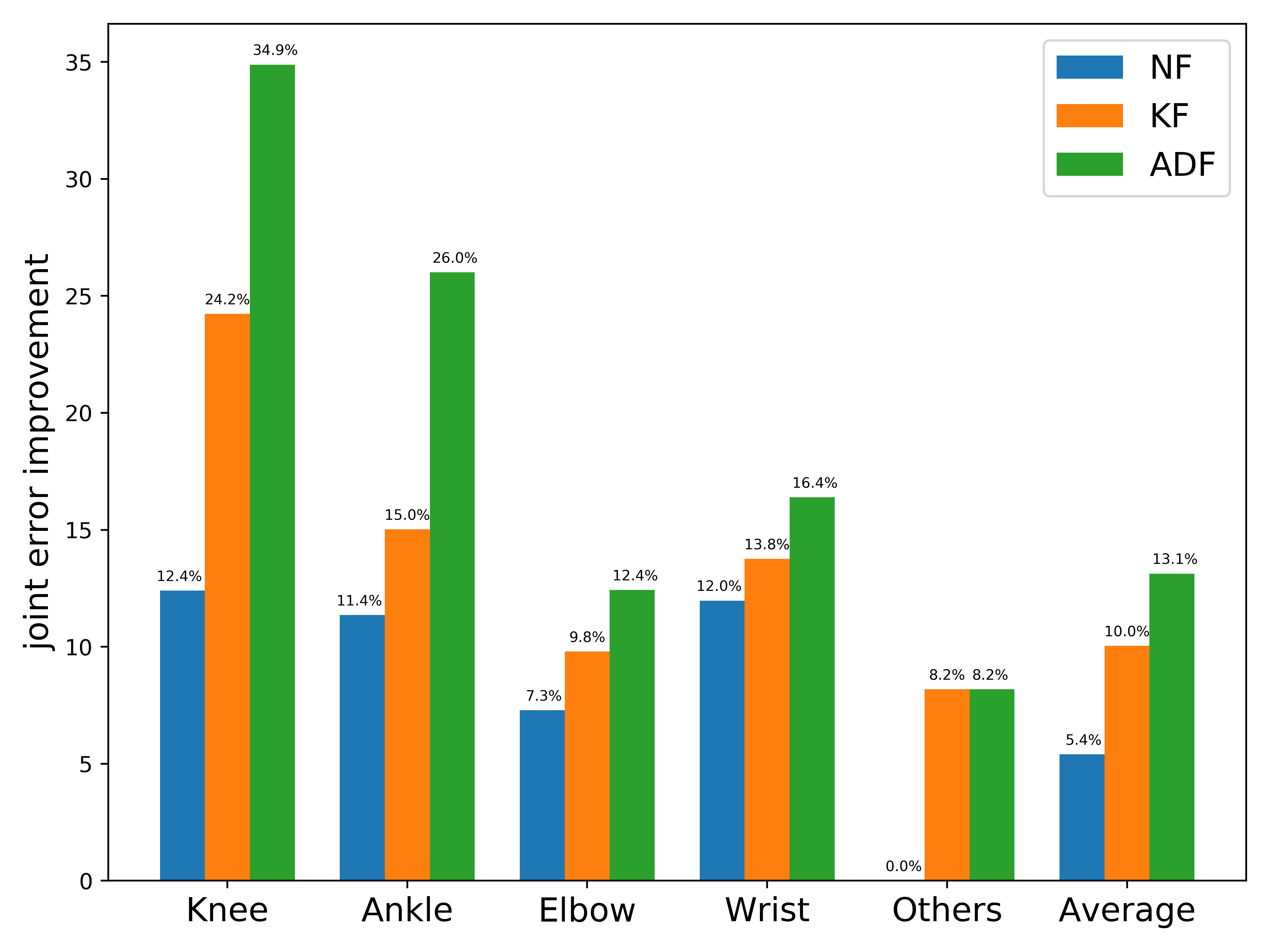}}
\caption{The respective joint error improvement of three proposed approaches. 'Others' represents joints that are not related to used IMUs.}
\label{fig5}
\end{center}
\end{figure}

~\\
\noindent \textbf{Fusion Methods.}
\quad In this work we propose a total of three IMU-vision fusion approaches. We perform ablation studies of all three approaches in Total Capture dataset to evaluate their efficiency. We calculate the joint errors on the estimated 3D pose $P^{vis}$ by AlgTri as the baseline. Then, we conduct experiments on NaiveFuse (NF), KineFuse (KF) and AdaDeepFuse (ADF) and calculate the joints error on the fused 3D pose $P^{NF}$, $P^{KF}$ and $P^{ADF}$, respectively. Table \ref{table:ablation of tc} lists the improvements the three methods bring in the considered joints. Moreover, we use the results of baseline to minus the results of the fused poses and obtain the respective joint error improvement of three approaches, which are shown in Figure \ref{fig5} for more obvious illustration. 

\begin{table*}[!tp]
\caption{3D pose estimation errors (mm) of the ablation study in Total Capture dataset.}
\begin{center}
\scalebox{0.75}{
\resizebox{\textwidth}{!}{
\begin{tabular}{ccccc}
\toprule
Joint Error (mm) & Baseline & Naive Fuse & Kinematic Fuse & Adaptive Deep Fuse\\
\midrule
Belly &9.9 &9.9 (-) &9.6 ( $\downarrow$ 0.3) &9.6 ($\downarrow$ 0.3)\\
Neck &26.2 &26.2 (-) &26.4 ($\uparrow$ 0.2)  &26.4 ($\uparrow$ 0.2) \\
Nose &28.5 &28.5 (-) &28.2 ($\downarrow$ 0.3)  &28.2 ($\downarrow$ 0.3) \\
Right Hip &11.2 &11.2 (-) &10.2 ($\downarrow$ 1.0)  &10.2 ($\downarrow$ 1.0) \\
Left Hip &10.2 &10.2 (-) &8.5 ($\downarrow$ 1.7)  &8.5 ($\downarrow$ 1.7) \\
Right Shoulder &36.9 &36.9 (-) &31.2 ($\downarrow$ 5.7) &31.2 ($\downarrow$ 5.7) \\
Left Shoulder &38.4 &38.4 (-) &34.0 ($\downarrow$ 4.4)  &34.0 ($\downarrow$ 4.4) \\
\midrule
Right Knee &25.3 &23.1 ($\downarrow$ 2.2) &19.1 ($\downarrow$ 6.2) &16.7 ($\downarrow$ 8.6)\\
Right Ankle &27.7 &25.2 ($\downarrow$ 2.5) &23.2 ($\downarrow$ 4.5) &20.5 ($\downarrow$ 7.2)\\
Left Knee &26.3 &22.1 ($\downarrow$ 4.2) &20.0 ($\downarrow$ 6.3) &16.8 ($\downarrow$ 9.5)\\
Left Ankle &26.9 &23.2 ($\downarrow$ 3.7) &23.2 ($\downarrow$ 3.7) &19.8 ($\downarrow$ 7.1)\\
Right Elbow &39.9 &38.1 ($\downarrow$ 1.7) &35.9 ($\downarrow$ 4.0) &35.0 ($\downarrow$ 4.9)\\
Right Wrist &45.8 &42.1 ($\downarrow$ 3.7) &40.6 ($\downarrow$ 5.2) &38.9 ($\downarrow$ 6.9)\\
Left Elbow &43.8 &39.5 ($\downarrow$ 4.3) &39.5 ($\downarrow$ 4.3) &38.3 ($\downarrow$ 5.5)\\
Left Wrist &48.4 &41.7 ($\downarrow$ 6.7) &41.4 ($\downarrow$ 7.0) &40.7 ($\downarrow$ 7.7)\\
\hline
Average &25.9 &24.5 ($\downarrow$ 1.4) &23.3 ($\downarrow$ 2.6) &22.5 ($\downarrow$ 3.4)\\
\bottomrule
\end{tabular}}}
\end{center}
\label{table:ablation of tc}
\end{table*}

\begin{figure*}[!tp]
 \begin{center}
 	\centerline{\includegraphics[width= \linewidth]{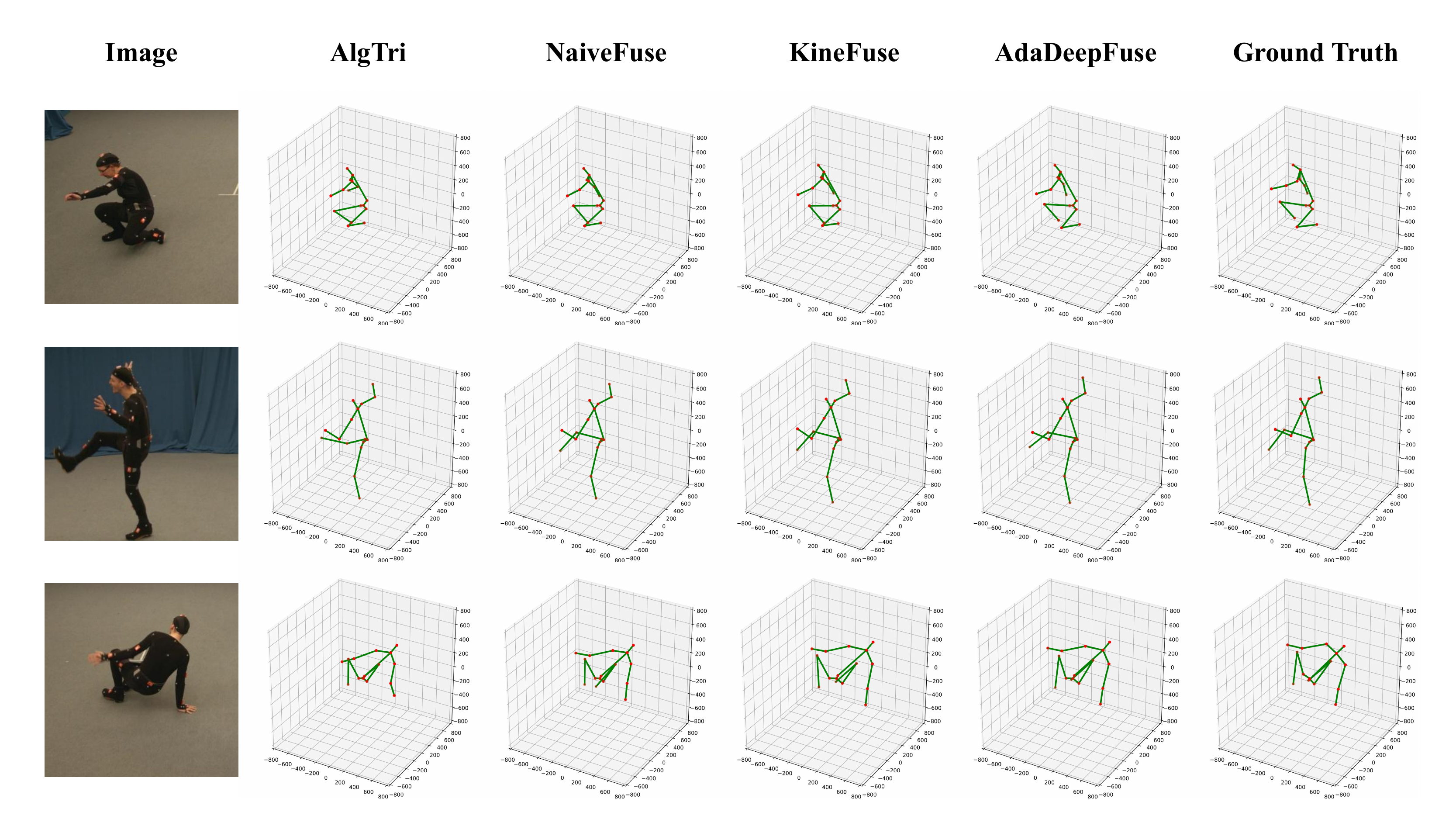}}
\caption{Illustration of different 3D pose results on Total Capture dataset. Self-occlusions exist in most of used image views on the selected cases, while being solved via kinematic constrains and efficient IMU-vision fusion.}
\label{fig6}
\end{center}
\end{figure*}

We can draw several conclusions by observing the results. First, we can find that NF only brings error improvement for the joints that are related to the used IMUs, while having no impact on other joints ($0.0\%$ of blue bar in 'Others'). This is the limitation of directly replacing local bone vectors while ignoring the holistic human skeleton. As a contrast, KF and ADF both prompt better performance on these IMU-unrelated joints ($8.3\%$ of orange and green bars in 'Others'). This is mainly because IK and FK ensure the bone lengths consistency and avoid the bone stretch issue in the fused 3D pose. This conclusion can be also achieved by observing Figure \ref{fig6}, where we draw the output 3D poses of different methods and show them. The bone lengths in the 3D pose of AlgTri and NF disagree with that of ground truth, while this phenomenon does not occur in the results of KF and ADF. Based on the more accurate ancestor joints, the errors of IMU-related joints on limbs are further improved in KF, compared to NF. The whole average joint error improvement is also increased from $5.4\%$ to $10.0\%$. The value is further boosted to $13.1\%$ in ADF, indicating the efficiency of the adaptive deep learning-based fusion strategy.

In order to figure out which IMU plays more important roles in the fusion process, we conduct experiments on the usage of IMUs based on KineFuse. The results are listed in Table \ref{table3}. It can be observed that the IMUs in upper limbs perform better than the IMUs in lower limbs. This is because the error improvement of the upstream joints will further correct the downstream children. In this work, using all IMUs in limbs serves as the final setting for the most accurate results.

\begin{table}[!tp]
\caption{Ablation study on the usage of IMUs. Upper represents the upper arms and thighs while lower represents the forearms and calves. 4 and 8 denotes the number of IMUs attached to body bones.}
\begin{center}
\scalebox{0.48}{
\resizebox{\textwidth}{!}{
\begin{tabular}{cccc}
\toprule
IMUs & upper limbs (4) & lower limbs (4) & limbs (8)\\
\midrule
MPJPE&25.2&25.7&23.3\\
\bottomrule
\end{tabular}}}
\end{center}
\label{table3}
\end{table}

\begin{figure*}[!tp]
 \begin{center}
 	\centerline{\includegraphics[width= 475pt]{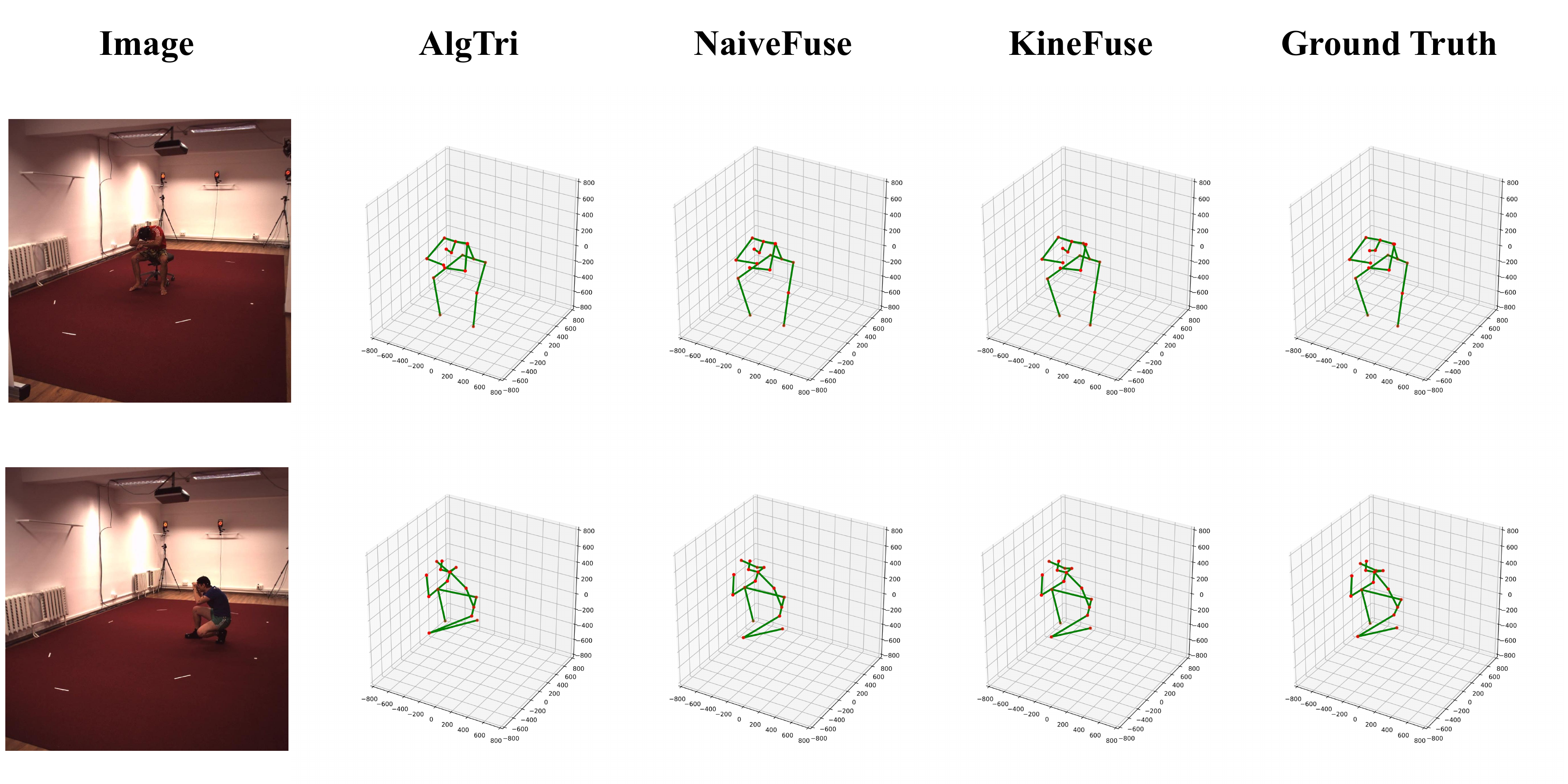}}
\caption{Illustration of different 3D pose results on Human3.6M dataset. Since there are no real IMU information, AdaDeepFuse is not explored here.}
\label{fig7}
\end{center}
\end{figure*}

We also conduct experiments on Human3.6M dataset for generalization evaluation. Due to the absolute accuracy of simulative IMU information, we just validate NaiveFuse and KineFuse on the testing set. The qualitative and quantitative results are shown in Table \ref{table4} and Figure \ref{fig7}. We can obtain similar conclusions with that from experiments on Total Capture. The IMU-related joint errors are improved from $30.1mm$
to $23.5mm$ by NaiveFuse, while to $19.2mm$ by KineFuse. In addition, results of other joints are also decreased by KineFuse from $20.4mm$ to $18.4mm$ while fixed in NaiveFuse. Also, in Figure \ref{fig7}, the bone length consistency is ensured in KineFuse while not in AlgTri and NaiveFuse.

\begin{table}[!tp]
\caption{3D pose estimation errors (mm) of the ablation study in Human3.6M dataset. IMU-related joints contain the knees, ankles, elbows and wrists.}
\begin{center}
\scalebox{0.48}{
\resizebox{\textwidth}{!}{
\begin{tabular}{lccc}
\toprule
Methods & IMU-related & Others& Average \\
\midrule
Baseline &30.1 &20.3 &24.9\\
NaiveFuse &23.5 ($\downarrow$ 6.6)& 20.3 (-)&21.8 ($\downarrow$ 3.1)\\
KineFuse &19.2 ($\downarrow$ 9.9)&18.4 ($\downarrow$ 1.9)&18.8 ($\downarrow$ 6.1)\\
\bottomrule
\end{tabular}}}
\end{center}
\label{table4}
\end{table}

\begin{figure}[!tp]
 \begin{center}
 	\centerline{\includegraphics[width= \linewidth]{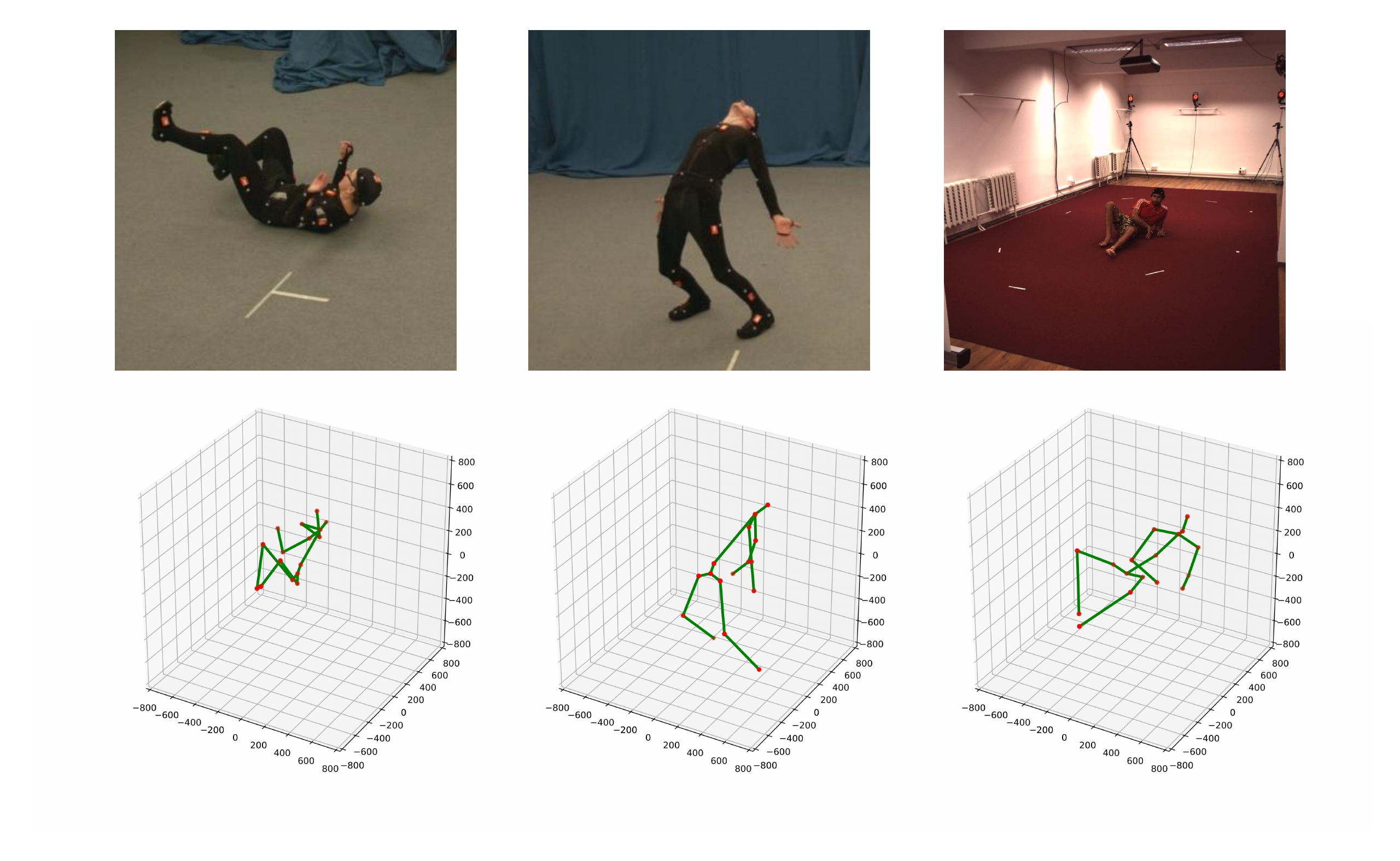}}
\caption{Illustration of some failure cases on Total Capture dataset (left and middle) and Human3.6M dataset (right). They are mainly extremely unusual poses. It is hard to estimate the upper stream joints such as belly, hips and shoulders in these cases.}
\label{fig8}
\end{center}
\end{figure}

\section{Analysis and Discussion}

As shown in Figure \ref{fig6} and \ref{fig7}, the baseline method based on AlgTri performs not well on the unusual actions or on the poses where occlusions exist. The IMU information is occlusion-free so the aligned bone vectors can help improve the results of limbs by NaiveFuse. However, due to the limitation of resolution of the estimated heatmaps, the reconstructed joint locations are variable. The human skeleton is with the encumbrance of bone stretch issue. In this work, we figure out this issue by applying IK and FK layers and thus ensuring the bone length consistency. Based on this improvement, the local rotation information is further utilized for better IMU-vision fusion. In KineFuse and AdaDeepFuse, we explore two effective approaches and achieve superior performance. 

We show some failure cases on two utilized datasets in Figure \ref{fig8}. The first two columns are samples from Total Capture using AdaDeepFuse, while the last column is from Human3.6M using KineFuse. The main reason for failure is that the predictions on the upstream joints are unsatisfactory, resulting in poorer predictions on the downstream joints under the usage of IMU information.

Although our approaches achieved the state-of-the-art performance, there still exists a lot of work to exploit in the future. First, in addition to being able to represent local rotation, another important advantage of IMU information locates in its temporal continuity and stability in fast motion. In this work, we transfer estimated joint locations to pose parameters The pose parameters, which can represent local rotations, also have an advantage on interpolation for pulse error compensation caused by occlusions on motions. The IMU-vision fusion based on temporal information is very valuable to explore in our future work. Another limitation of the proposed approaches is that they rely on the known lengths to build the rest T-pose, which is hard to receive in the real application. Thus, we will try to integrate the prediction of bone lengths into the whole framework.

\section{Conclusion}

We propose a framework for IMU-vision fusion based on a parametric human representation. The information from IMU and vision data is aligned as bone vectors or local rotations and is aggregated via three effective approaches. The integrated IMU information not only corrects offset error of the estimated joints caused by occlusions and quantization error on heatmaps, but also aids to yield a plausible and authentic human pose. Extensive experiments with ablation analysis show that our proposed framework achieves superior performance on Total Capture dataset. Also, our approaches are proved able to generalize to other benchmarks via validation experiments on Human3.6M dataset. The value of temporal continuity in IMU-vision fusion is going to be explored in the future.

{\small
\bibliographystyle{ieee_fullname}
\bibliography{egbib}
}








\end{document}